\documentclass[10pt,twocolumn,letterpaper]{article}

\usepackage{multirow}
\usepackage{graphicx}
\usepackage{bm}
\usepackage[pagenumbers]{cvpr} %

\newcommand\coolname{\textcolor{black}{\sc\textbf{Ctrl-D}}\xspace}

\usepackage{hyperref}
\usepackage{lipsum}

\usepackage[capitalize,nameinlink]{cleveref} %

\usepackage{abbreviations}

\definecolor{cvprblue}{rgb}{0.21,0.49,0.74}

\hypersetup{ %
    breaklinks,
    colorlinks,
    allcolors=cvprblue
}

\def\paperID{****} %
\def\confName{CVPR}
\def\confYear{2025}

\title{\hspace{-0.7mm}\coolname: Controllable Dynamic 3D Scene Editing with Personalized 2D Diffusion}

\author{Kai He$^{1,2}$ \quad \qquad Chin-Hsuan Wu$^{1,2}$ \quad \qquad Igor Gilitschenski$^{1,2}$\\
$^{1}$University of Toronto \quad $^{2}$Vector Institute
}

\begin{document}

\twocolumn[{
\renewcommand\twocolumn[1][]{#1}
\maketitle
\vspace{-1.2cm}
\begin{center}
    \centering
    \vspace{5pt}
    \captionsetup{type=figure}
    \includegraphics[clip, trim=0cm 7.7cm 3.5cm 0cm, width=1.0\textwidth]{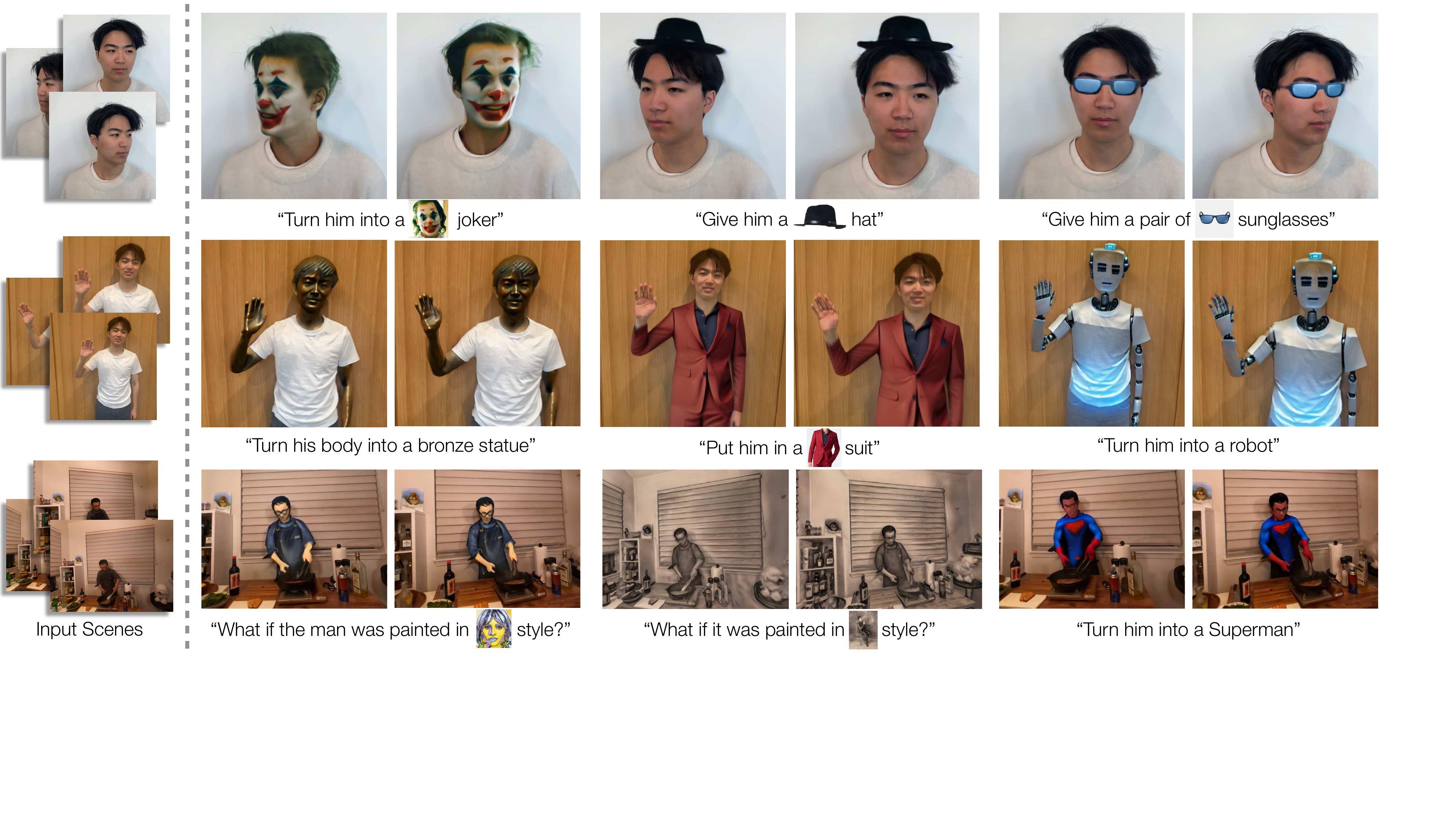}
    \vspace*{-0.7cm}
    \captionof{figure}{We present \coolname, a dynamic 3D scene editing framework that enables controllable, high-quality, consistent scene edits by editing only a single image using any 2D editing approach. Our framework is also compatible with both monocular and multi-camera scenes. \textbf{Please refer to our project page for dynamic visualizations.}}
    \label{fig:teaser}
\end{center}
}]

\begin{abstract}

\vspace{-0.4cm}
Recent advances in 3D representations, such as Neural Radiance Fields and 3D Gaussian Splatting, have greatly improved realistic scene modeling and novel-view synthesis. However, achieving controllable and consistent editing in dynamic 3D scenes remains a significant challenge. Previous work is largely constrained by its editing backbones, resulting in inconsistent edits and limited controllability. In our work, we introduce a novel framework that first fine-tunes the InstructPix2Pix model, followed by a two-stage optimization of the scene based on deformable 3D Gaussians. Our fine-tuning enables the model to ``learn'' the editing ability from a single edited reference image, transforming the complex task of dynamic scene editing into a simple 2D image editing process. By directly learning editing regions and styles from the reference, our approach enables consistent and precise local edits without the need for tracking desired editing regions, effectively addressing key challenges in dynamic scene editing. Then, our two-stage optimization progressively edits the trained dynamic scene, using a designed edited image buffer to accelerate convergence and improve temporal consistency. Compared to state-of-the-art methods, our approach offers more flexible and controllable local scene editing, achieving high-quality and consistent results. Please see our website for more details: \textcolor{magenta}{\href{https://IHe-KaiI.github.io/CTRL-D/}{https://IHe-KaiI.github.io/CTRL-D/}}.

\end{abstract}
    
\vspace{-0.5cm}
\section{Introduction}
\label{sec:intro}

The advent of efficient 3D representations, such as Neural Radiance Fields (NeRF)~\cite{nerf} and 3D Gaussian Splatting (3DGS)~\cite{3dgs}, has revolutionized the real-world scene capture, enabling photorealistic novel-view synthesis. With a growing interest in dynamic scenes, several works~\cite{nerfplayer, cao2023hexplane, yang2024deformable, yang2023gs4d, Wu_2024_CVPR} have extended NeRF and 3DGS to model such scenes, significantly broadening their scope of applications. Building on these developments, there is an increasing demand for editing existing scenes—a capability essential for various applications such as virtual and augmented reality (VR/AR), data augmentation, and content creation. While recent advances in static 3D scene editing have achieved promising results, the controllable and consistent editing of dynamic 3D scenes remains underexplored.

Recent progress~\cite{instruct_4d24d, control4d} in dynamic 3D scene editing has leveraged pre-trained diffusion-based image editing methods, such as InstructPix2Pix (IP2P)~\cite{instructpix2pix} and ControlNet~\cite{controlnet}, for iteratively editing the scene. However, these approaches often face limitations due to the constrained editing capacity of their backbones, such as the inability to perform precise local editing or restrictions imposed by the training domain of the pre-trained diffusion models. This leads to inconsistent editing results across frames or limited controllability. Overcoming these limitations for dynamic scene editing is an open research challenge. Inspired by recent strides in 2D editing~\cite{instructpix2pix, controlnet, sdedit}, it has become increasingly robust and versatile. We investigate how a single edited image can serve as a reference for dynamic scene editing. This approach offers greater flexibility and controllability. It allows users to employ any 2D editing tool to modify the image, and then guides edits in dynamic scenes.

In this paper, we propose \coolname, a novel pipeline that first fine-tunes the IP2P model using only one edited image, enabling personalized dynamic scene editing with expanded editing capabilities. This personalization enables IP2P to ``learn'' the editing ability from the given reference image. Controllable local editing in dynamic scenes is challenging, as tracking the desired editing region is more difficult than in static 3D scenes. Previous method~\cite{watch_your_steps} relies on noise differences to determine editing regions. In contrast, our personalized IP2P directly learns these editing regions from the reference image. This enables it to focus on relevant areas during localized editing, thereby enhancing stability and practical usability. Furthermore, we incorporate a prior preservation loss inspired by~\cite{dreambooth}. It enables our personalized IP2P to maintain paired-edit consistency and stability.

After fine-tuning, the personalized IP2P can be deployed for dynamic scene editing tasks. We then introduce a two-stage training framework based on deformable 3D Gaussians~\cite{yang2024deformable, Wu_2024_CVPR}, designed to improve efficiency in dynamic scene editing. In the first stage, only the canonical space of the scene is optimized, with Gaussian densification performed based on a single keyframe. In the subsequent stage, both the deformation field and 3D Gaussians are optimized with a specially designed edited image buffer, which accelerates the scene to converge during optimization. This framework significantly enhances the temporal consistency of editing and outperforms existing state-of-the-art methods for dynamic scene editing~\cite{instruct_4d24d}.

To summarize, our contributions are threefold: 
\begin{itemize} 
    \item We introduce a novel scene editing framework that fine-tunes IP2P using a reference edited image, expanding editing capabilities while preserving the consistency of the original model's editing style. 
    \item We present a two-stage dynamic scene optimization based on deformable 3D Gaussians, enabling efficient and consistent dynamic scene editing. 
    \item Our approach transforms the complex dynamic scene editing task into a simple 2D image editing, making dynamic scene editing more accessible and controllable, enabling consistent local editing, and showing superior results than previous work.
\end{itemize}

\section{Related Work}
\label{sec:related}

\begin{figure*}[ht]
  \centering
  \includegraphics[clip, trim=0cm 21.1cm 5.3cm 0cm, width=1.0\textwidth]{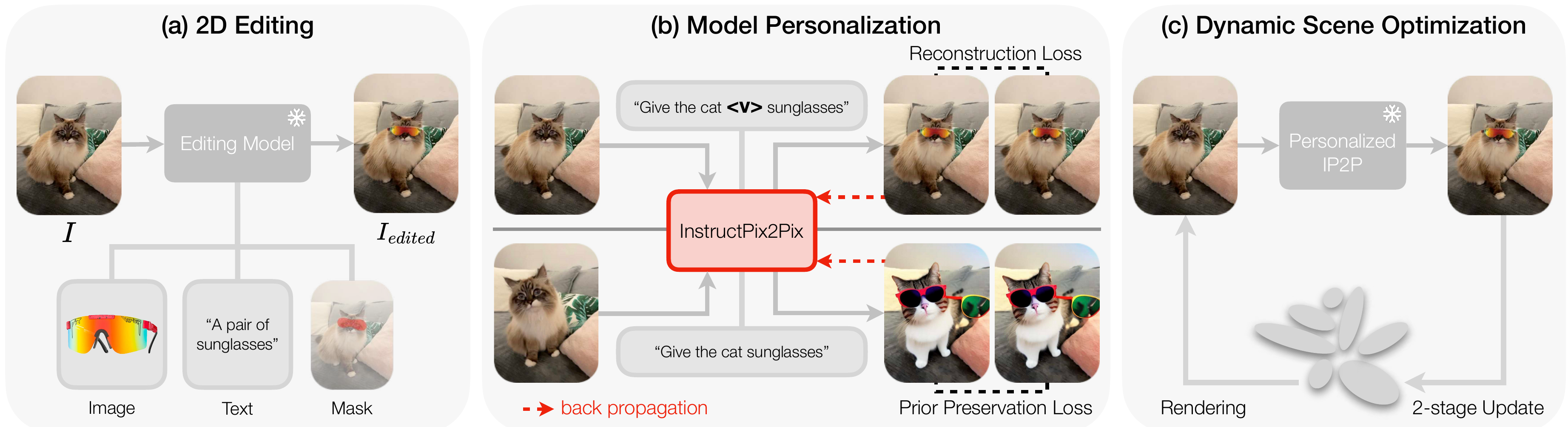}
  \vspace{-6mm}
  \caption{Our pipeline for controllable dynamic scene editing. Given a dynamic 3D scene, our method (a) first edits one frame as a reference with any 2D editing model, we then (b) fine-tune the InstructPix2Pix \cite{instructpix2pix} with the edited reference image, along with sampled images from the original models to preserve the model priors, and then (c) we optimize the dynamic 3D scenes with deformable gaussian representation, using the designed 2-stage method.}
  \label{fig:pipeline}
\end{figure*}

\paragraph{Diffusion Models for 2D Editing.}

Diffusion models~\cite{ddpm, score, nonequilibrium} have emerged as the mainstream approach for image synthesis owing to their remarkable ability to generate high-quality images. By training these models on large-scale datasets, many works have demonstrated impressive results in Text-to-Image (T2I) generation~\cite{stable_diffusion, glide, dalle_2, imagen}, creating realistic images from diverse text descriptions. Building on this success, pre-trained T2I diffusion models have also been adapted for image editing via text prompts~\cite{imagic, sdedit, blended, prompt2prompt, diffedit, textual_inversion, dreambooth, repaint}. Notably, SDEdit~\cite{sdedit} modifies the input image by adding and removing noise based on the target prompt. Textual Inversion~\cite{textual_inversion} and DreamBooth~\cite{dreambooth} enable personalized generation by learning the concept of a specific object from a small collection of images. Imagic~\cite{dreambooth} refines a single image by inverting and fine-tuning it, then regenerates the image with a new text prompt. Another line of work performs local inpainting, guided by both a caption and an additional region constraint, which can be either model-generated~\cite{diffedit, prompt2prompt} or user-defined~\cite{repaint, blended}. InstructPix2Pix (IP2P) ~\cite{instructpix2pix} further leverages Prompt-to-Prompt~\cite{prompt2prompt} to collect image editing examples for training, achieving superior quality and speed compared to previous methods. ControlNet~\cite{controlnet}, by contrast, is not specifically focused on editing but rather extends T2I models with multiple input conditions, effectively enhancing editing capabilities. Recently, video editing has become an area of increased focus. Early works, like Tune-A-Video~\cite{tune_a_video} and Video-P2P~\cite{video_p2p}, enable localized edits by fine-tuning T2I models to capture continuous motion. While many follow-up methods~\cite{pix2video, controlvideo, videoswap, ku2024anyv2v, chen2024unictrl} achieve plausible results, they still face challenges in view and temporal consistency, and quality generation.

\paragraph{NeRF and 3DGS Editing.}

Neural Radiance Field (NeRF)~\cite{nerf} and 3D Gaussian Splatting (3DGS)~\cite{3dgs} have revolutionized 3D modeling with their high-fidelity rendering quality. This success has naturally led to the development of tools for editing both NeRF and 3DGS. Beginning with EditNeRF~\cite{editnerf}, which optimizes latent codes to refine both the appearance and shape of objects, subsequent works~\cite{clip_nerf, textdeformer, sine} have expanded editing capabilities by incorporating text prompts, guided by CLIP~\cite{clip} for optimization. Other approaches enable localized editing by applying inpainting techniques from 2D manipulation to gather target datasets and update the learned fields~\cite{kobayashi2022decomposing, spin_nerf}. Additionally, some methods utilize predefined template models (e.g., meshes~\cite{neural_body, xu2022deforming, neumesh}, skeletons~\cite{neural_articulated}, or point clouds~\cite{neuraleditor}) to articulate and deform shapes within the radiance field.

Recently, with the rapid advancement of T2I diffusion and video generation models, leveraging these priors for 3D~\cite{instruct_nerf2nerf, vox_e, dreameditor, gaussianeditor, watch_your_steps, customize_your_nerf} and 4D~\cite{instruct_4d24d, control4d} editing has gained substantial attention. While Instruct 4D-to-4D~\cite{instruct_4d24d}, Control4D~\cite{control4d} share our objective of achieving effective dynamic editing, they often produce inconsistent edits across frames and offer limited control over the scene.

\paragraph{Dynamic View Synthesis.}

Building on the success of NeRF, numerous works have extended it to model dynamic scenes. Some approaches~\cite{gao2021dynamic, gao2022dynamic, dyNeRF, dct_NeRF, videoNeRF} incorporate time as an additional dimension in the radiance field to capture scene deformations. However, these methods often struggle to model complex geometry, leading to unrealistic, inconsistent structures in occluded regions. Another class of methods~\cite{nerfies, hypernerf, tretschk2021nonrigid, d_nerf} introduces a deformation field to disentangle time from the radiance field, where deformations, conditioned on time~\cite{d_nerf} or a temporal latent code~\cite{nerfies, hypernerf, tretschk2021nonrigid}, are parameterized as rigid body motion. While effective for longer videos, these approaches generally perform best on object-centric scenes with minimal motion or simple camera paths. Recently, the introduction of 3DGS has advanced this field by providing a lightweight, explicit deformation model to handle Gaussian motions and shape changes~\cite{yang2023gs4d, luiten2023dynamic, yang2024deformable}, Despite the promising quality of dynamic GS models, methods for editing such representations are still largely unexplored.

\section{Method}
\label{sec:method}

Our method takes as input a trained dynamic 3D Gaussian scene, along with the original source data: a set of multi-view images and their registered camera poses. The framework consists of three key modules: First, we select a single image from the input set and apply any off-the-shelf image editing technique to produce an edited version of this image (\cref{sec:singleImageEditing}). Using the pair of source and edited images, we then fine-tune the InstructPix2Pix (IP2P) model with a specifically designed text prompt (\cref{sec:personalization}). Finally, we introduce a two-stage optimization process for the dynamic 3D Gaussian model, which iteratively refines the dynamic scene using the personalized image editor (\cref{sec:optimization}). An overview of the pipeline is shown in Fig.~\ref{fig:pipeline}.

\subsection{Background}
\paragraph{Dynamic 3D Gaussian Splatting.}

3D Gaussian Splatting (3DGS)~\cite{3dgs} utilizes a set of 3D Gaussians $G=(\boldsymbol{x}, \boldsymbol{r}, \boldsymbol{s}, \sigma)$ defined by a center position $\boldsymbol{x}$, opacity $\sigma$, and 3D covariance matrix $\Sigma$ obtained from quaternion $\boldsymbol{r}$ and scaling $\boldsymbol{s}$. The color of the pixel on the image plane, denoted by $\mathbf{p}$, is rendered using a differentiable splatting rendering process as follows:
\begin{equation}
  \begin{aligned} 
    C(\mathbf{p}) & =\sum_{i \in N} \alpha_i c_i \prod_{j = 1}^{i - 1} (1 - \alpha_j), \\
    \alpha_i & =\sigma_i \mathrm{e}^{-\frac{1}{2}\left(\mathbf{p}-\bm{\mu}_i\right)^T \sum^{\prime}\left(\mathbf{p}-\bm{\mu}_i\right)},
  \end{aligned}
\end{equation}
where $c_i$ represents the color of each Gaussian along the ray, $\Sigma'$  is the corresponding 2D covariance matrix, and $\bm{\mu}_i$ represents the $uv$ coordinates of the 3D Gaussians projected onto the 2D image plane.

To model dynamic scenes, some works~\cite{yang2024deformable, Wu_2024_CVPR} extend 3DGS with a deformation field, enabling the learning of 3D Gaussians in canonical space, and subsequently puts the deformed 3D Gaussians $G=(\boldsymbol{x} + \delta \boldsymbol{x}, \boldsymbol{r} + \delta \boldsymbol{x}, \boldsymbol{s} + \delta \boldsymbol{x}, \sigma)$ into the rasterization pipeline. Given the time $t$ and center position $\boldsymbol{x}$ of 3D Gaussians as inputs, the deformation MLP predicts offsets:
\begin{equation}
(\delta \boldsymbol{x}, \delta \boldsymbol{r}, \delta \boldsymbol{s})=\mathcal{F}_\theta(\gamma(\operatorname{sg}(\boldsymbol{x})), \gamma(t)),
\end{equation}
where $\operatorname{sg}(\cdot)$ indicates a stop-gradient operation, and $\gamma$ represents the positional encoding.

\paragraph{InstructPix2Pix.}

InstructPix2Pix (IP2P) \cite{instructpix2pix} is a diffusion-based method for 2D image editing. Given an image $I$ and a text instruction $C_T$, IP2P aims to generate an edited image $z_0$ based on $I$ following the instruction guidance of $C_T$. The diffusion model predicts noise in the input image (or pure noise) $z_t$ at timestep $t$, using a denoising U-Net $\epsilon_\theta$. IP2P is trained on a dataset in which, for each pair $I$ and $C_T$, an example edited image $I_{\mathrm{edited}}$ is provided. The method builds on latent diffusion \cite{stable_diffusion}, incorporating a variational autoencoder \cite{kingma2013auto} (VAE) with an encoder $\mathcal{E}$ and a decoder $\mathcal{D}$. During training, Gaussian noise $\epsilon \sim \mathcal{N}(0, 1)$ is added to $z = \mathcal{E}(I_{\text{edited}})$ to obtain a noisy latent representation $z_t$, with $t \in T$ as a randomly selected timestep. The denoiser $\epsilon_\theta$ is initialized with weights from stable diffusion \cite{stable_diffusion} and fine-tuned to minimize the diffusion objective:
\begin{equation}
\mathbb{E}_{I_{\mathrm{edited}}, I, C_T, \epsilon, t} \left[ \| \epsilon - \epsilon_\theta (z_t, t, I, C_T)\|_2 ^ 2\right].
\end{equation}
To achieve classifier-free guidance for both image conditioning and text conditioning, IP2P randomly sets $I = \varnothing_I$ for 5\% of examples, $C_T = \varnothing_T$ for 5\% of examples, and both $I = \varnothing_I$ and $C_T = \varnothing_T$ for 5\% of examples during training. Thus, we can control the strength of the edit by the image guidance scale $s_I$ and the text guidance scale $s_T$. The modified score estimate is then computed as
\begin{equation}
  \begin{aligned} 
    \tilde{e_\theta}\left(z_t, t, I, C_T\right)& = e_\theta\left(z_t, t, \varnothing_I, \varnothing_T \right) \\ +& s_I \left(e_\theta\left(z_t, t, I, \varnothing_T\right)-e_\theta\left(z_t, t, \varnothing_I, \varnothing_T\right)\right) \\ +& s_T \left(e_\theta\left(z_t, t, I, C_T\right)-e_\theta\left(z_t, t, I, \varnothing_T\right)\right).
  \end{aligned}
\end{equation}

\subsection{Personalization of InstructPix2Pix}
\label{sec:personalization}

Assuming we have obtained the edited image $I_{\mathrm{edited}}$ and the source image $I$ from the previous stage, we proceed with fine-tuning the U-Net $\epsilon_\theta$ of InstructPix2Pix (IP2P) to create a personalized editing model. This fine-tuning process allows the model to specialize in our desired edits while preserving its prior capabilities. To achieve this, we require a text instruction that accurately describes the editing action. Initially, we utilize GPT-4V \cite{achiam2023gpt} to generate a preliminary text instruction $C_T^\star$ to describe the edit. To enhance specificity, we follow the approach in~\cite{dreameditor}, incorporating a specialized token $\texttt{<V>}$ before the last noun or adjective describing the editing target. This process results in a specific text prompt $C_T$, tailored to describe our editing process.

To retain the generalization capability of IP2P, we integrate a prior preservation loss inspired by DreamBooth~\cite{dreambooth}. This approach helps maintain the model's adaptability while focusing on specific edits. The input for prior preservation training includes the initial text instruction $C_T ^ \star$, a randomly selected frame $I_d$ from the dynamic scene dataset, and a corresponding edited image $I_{\mathrm{edited}} ^ \star$ generated by the original IP2P with text condition $C_T ^ \star$ and image condition $I_d$. This configuration enables the model to learn new edits while preserving its prior knowledge.

The fine-tuning loss function for our method consists of two terms. The primary term minimizes the error between the predicted noise and the actual noise in the target-edited image, while the prior preservation term ensures that the model does not lose generalization over the original editing capabilities. The overall finetuning loss is defined as:

\begin{equation}
    \begin{aligned}
    \mathcal{L}_{\mathrm{finetune}} = &\mathbb{E}_{I_{\mathrm{edited}}, I, C_T, \epsilon, t} \left[ \| \epsilon - \epsilon_\theta (z_t, t, I, C_T)\|_2 ^ 2\right] + \\
    \lambda &\mathbb{E}_{I_{\mathrm{edited}} ^ \star, I_d, C_T ^ \star, \epsilon, t} \left[ \| \epsilon - \epsilon_\theta (z_t ^ \star, t, I_d, C_T ^ \star)\|_2 ^ 2\right],
    \end{aligned}
\end{equation}
where $z_t$ denotes the noisy latent representation at timestep \( t \) extracted from the edited input images, and $z_t ^ \star$ denotes the corresponding variables used in prior preservation training. $\lambda$ is a hyperparameter that controls the relative weight of the prior-preservation term.

\paragraph{Data augmentation.}

In DreamBooth, fine-tuning typically requires 3–5 images, which allows for more robust model adaptation. However, in our setting, obtaining multiple paired images that achieve the same editing effect is challenging and often impractical. With only a single image pair available for fine-tuning, one-shot fine-tuning can cause overfitting and potential model collapse. Although various methods \cite{dong2022dreamartist, lu2024object} have been proposed to address these issues, they are not suitable for our editing framework. To overcome this limitation, we employ a straightforward and effective data augmentation technique. By applying affine transformations—including rotation, translation, and shear—to both the source and edited images, we find that it significantly reduces overfitting and enhances model stability during fine-tuning.

\subsection{Optimization of Dynamic 3D Gaussians}
\label{sec:optimization}

In our pipeline, we propose a two-stage optimization approach for dynamic 3D Gaussians to progressively edit the dynamic scene. First, we use the edited image from the single image editing stage as a keyframe to optimize the canonical space of the scene, while simultaneously performing Gaussian densification. Next, we optimize both the deformation field and the 3D Gaussians using a carefully designed edited image buffer, which allows for efficient and consistent editing results across frames.

\paragraph{Stage 1: Keyframe-guided Gaussian densification.} 

With the edited 2D image as a reference keyframe, we begin by optimizing the canonical space of the dynamic scene. In this initial stage, we refine only the 3D Gaussians, keeping the deformation field frozen. Unlike NeRF-based methods, 3DGS-based methods use point-based rendering. This discrete scene representation can lead to empty areas in the geometry of our edited regions, where Gaussians are sparse or absent, limiting the scene representation capability. The original 3DGS introduces adaptive control to densify the Gaussians for regions under reconstruction, while also subdividing areas where Gaussians are large and exhibit significant overlap. In our pipeline, we employ a similar adaptive scheme, using positional gradients to identify areas for densification. By optimizing the 3D Gaussians with only the edited keyframe and increasing density based on gradient statistics, we enhance the 3D Gaussians' capacity to express complex appearances and geometries for editing. This process also accelerates the overall optimization, providing a coarse scene for the next stage.

\paragraph{Stage 2: Optimization on both deformation fields and 3D Gaussians.}

In this stage, we fix the number of Gaussians and optimize both the deformation field and the 3D Gaussians. Following the idea of Instruct-NeRF2NeRF \cite{instruct_nerf2nerf}, we apply a similar process with the Iterative Dataset Update (Iterative DU) which iteratively replaces the dataset image with its edited counterpart. Benefiting from the relative consistency of our editing results across frames, we introduce an edited image buffer to accelerate the editing process. In each iteration, we randomly select a frame that will not be edited, and then use our personalized IP2P to generate the corresponding edited image, which is then added to the edited image buffer. Meanwhile, the 3D Gaussians and deformation field are trained solely on the images within this edited image buffer. We refer to this as the warm-up phase. Once every image has been edited at least once, resulting in an edited image buffer that contains all images, our strategy aligns with the original Iterative DU approach. Initially, only the keyframe is included in the edited image buffer.

After the warm-up phase, the preliminary results demonstrate reasonable temporal consistency. To further enhance this temporal consistency, we introduce a temporal loss $\mathcal{L}_{\mathrm{temp}}$ inspired by \cite{wang2024v}, which minimizes the differences between adjacent frames. The overall loss function is then expressed as:
\begin{equation}
    \mathcal{L} = (1 - \lambda_{d}) \mathcal{L}_1 + \lambda_{d} \mathcal{L}_{\mathrm{D-SSIM}} + \lambda_t \mathcal{L}_{\mathrm{temp}},
\end{equation}
where the first two terms are adapted from 3DGS, and \( \lambda_t \) represents the weight associated with our temporal regularization term.

\begin{figure*}[t]
  \centering
  \includegraphics[clip, trim=0cm 1.2cm 11.3cm 0cm, width=0.95\textwidth]{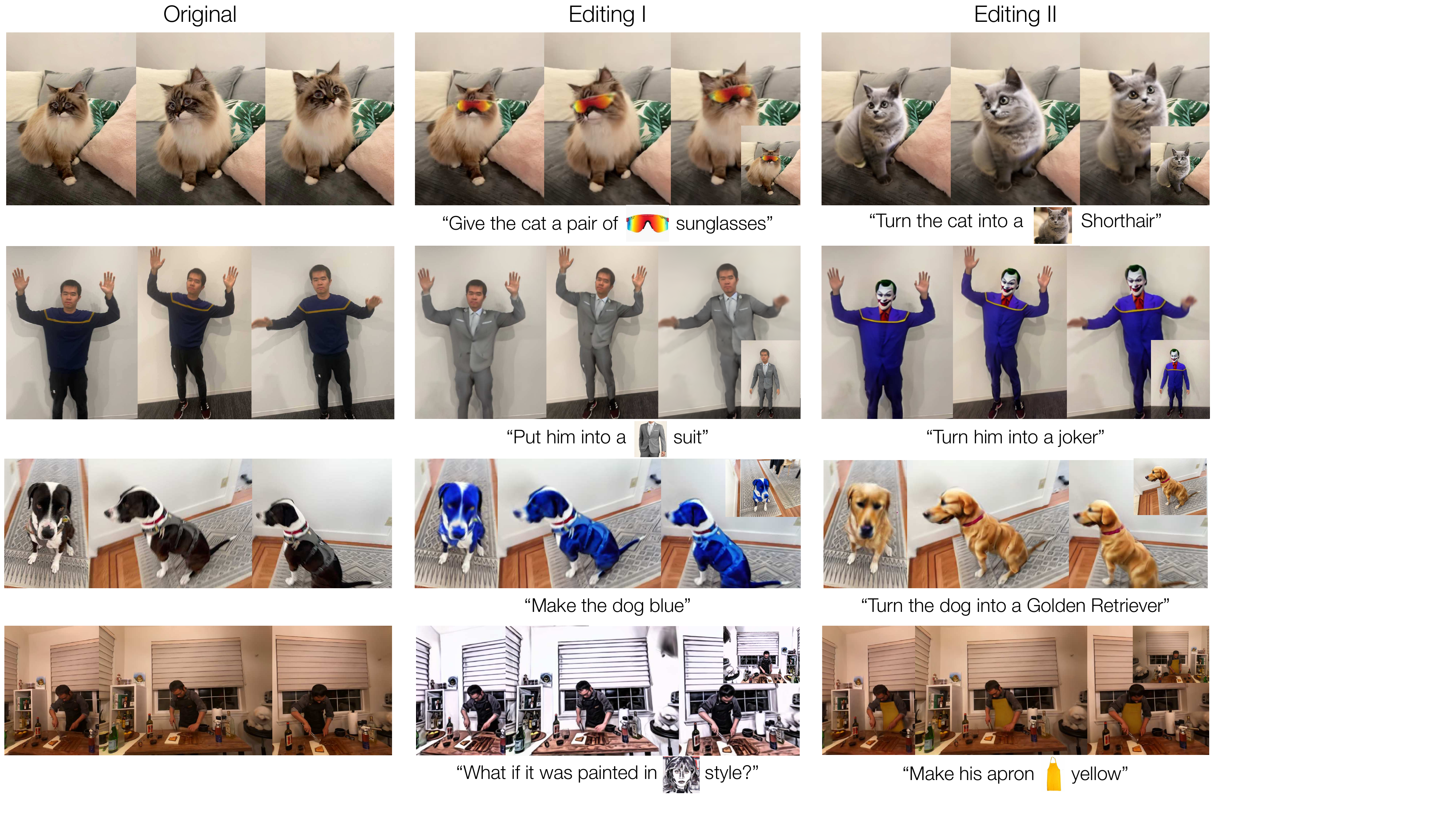}
  \vspace{-3mm}
  \caption{Qualitative results on both monocular and multi-camera scenes. For each scene, we show two edited versions based on the original, using various 2D editing techniques to demonstrate the high fidelity, quality, and controllability of our method. The reference 2D image for each edit appears in the bottom-right or bottom-left corner.}
  \label{fig:our_results}
\end{figure*}

\subsection{Implementation Details}

\paragraph{Single image editing.} 
\label{sec:singleImageEditing}
For text-driven editing, we use a text prompt along with an optional mask to guide the edit. We use SDEdit \cite{sdedit} or InstructPix2Pix \cite{instructpix2pix} to perform 2D editing. To constrain edits to specific regions, we incorporate BlendedDiffusion \cite{blended}, enabling precise control within masked areas when a mask is provided. In image-driven editing, we utilize a reference image along with a mask to perform editing. Following the approach outlined in MimicBrush \cite{chen2024MimicBrush}, we apply reference imitation to edit the image, ensuring that the edited output closely aligns with the content-specific attributes of the reference image. For stylization, we utilize NNST \cite{kolkin2022neural} to transfer style from a given reference style image.

\paragraph{Scene Optimization.} We edit the image from the dataset for every $50$ iterations. For the monocular scene, we adopt~\cite{yang2024deformable} to model the scene and perform stage 1 for the first $300$ iterations; for the multi-camera scene, we utilize~\cite{Wu_2024_CVPR} to model the scene and perform stage 1 for the first $100$ iterations. We set $\lambda_d = 0.2$ and $\lambda_t = 0.001$. Further details will be discussed in the appendix.

\section{Experiments}
\label{sec:exp}

\paragraph{Dataset.}

We evaluate our methods using several dynamic scenes from DyCheck~\cite{gao2022dynamic}, which consist of monocular and object-centric scenes, and N3DV~\cite{li2022neural}, which includes dynamic indoor scenes with a multi-camera setup. To more effectively evaluate the local editing capability of our methods, we further use the same approach as DyCheck to capture human-centric dynamic scenes, similar to data shown in Instruct-NeRF2NeRF~\cite{instruct_nerf2nerf}, including face-forward portraits and full-body scenes.

\begin{figure*}[t]
  \centering
  \includegraphics[clip, trim=0cm 21.5cm 41cm 0cm, width=0.95\textwidth]{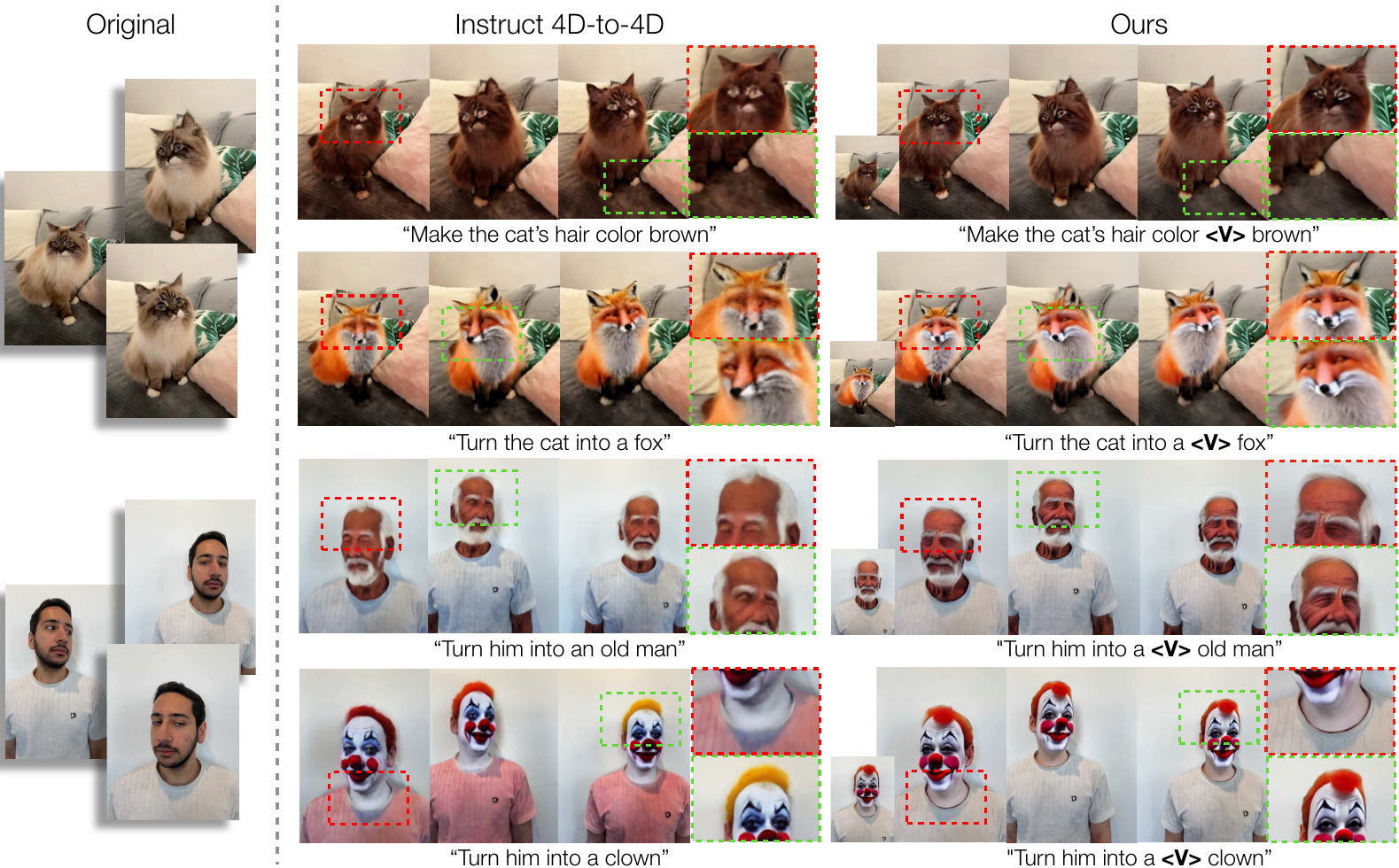}
  \vspace{-2mm}
  \caption{Qualitative comparison with Instruct 4D-to-4D (IN4D)~\cite{instruct_4d24d} on text-driven scene editing. The reference 2D images used in our method are shown at the bottom-left of each edited scene. We also provide zoomed-in details of the edited scene on the right side. Our method demonstrates superior consistency, higher quality, and more precise local edits, which are not achievable with IN4D.}
  \label{fig:comparison}
\end{figure*}

\subsection{Qualitative Evaluation}

Our qualitative results are presented in Fig.~\ref{fig:our_results}, where we evaluate our method on both monocular and multi-camera scenes. We apply various 2D editing techniques to the reference image, which is displayed in the bottom-right or bottom-left of each example. The results include text-prompt editing, image-prompt editing, and style transfer, all demonstrating high fidelity, quality, and controllability. Our method achieves strong temporal consistency across all scenes and multi-view consistency in multi-camera setups. Additionally, our method effectively supports local editing, preserving unrelated regions without modification. 

\begin{figure*}[t]
  \centering
  \includegraphics[clip, trim=0cm 27.5cm 0cm 0cm, width=1.0\textwidth]{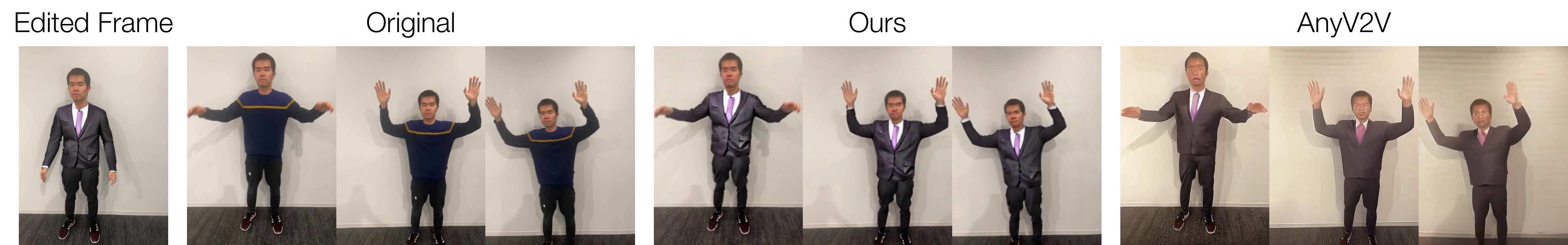}
  \caption{Qualitative comparison with AnyV2V~\cite{ku2024anyv2v} on monocular scenes. The leftmost image is the edited first frame. The prompt for our fine-tuned IP2P is ``Put him in a $\texttt{<V>}$ suit''. Our results demonstrate higher quality and greater consistency.}
  \label{fig:comparison_video}
\end{figure*}

\vspace{-3mm}

\paragraph{Comparison with dynamic scene editing methods.}

Two recent works, Instruct 4D-to-4D (IN4D)~\cite{instruct_4d24d} and Control4D~\cite{control4d}, focus on text-driven dynamic scene editing. Since Control4D is primarily tailored for portrait editing and has not released its code, we use Instruct 4D-to-4D as the baseline. Although our method supports both text-driven and image-driven editing with various single-image editing techniques, we limit the single-image editing approach to IP2P~\cite{instructpix2pix} with a mask to ensure fairness. The qualitative results are presented in Fig.~\ref{fig:comparison}. As shown, our method enables precise local editing, avoiding modifications to unrelated regions. For instance, in the first row, while IN4D alters the background and sofa colors, our approach modifies only the cat’s color, as intended. In the second and third rows, our method demonstrates superior temporal consistency across all frames, as well as enhanced quality. In the last row, IN4D affects the T-shirt color, but our method restricts changes to the face, matching the reference edited image. Additionally, our approach significantly enhances temporal consistency; in the last row, IN4D produces inconsistent hair colors across frames, while our results maintain color consistency. Lastly, our close-up images clearly show that our method achieves higher quality and fewer artifacts.

\begin{table}
\fontsize{8.4}{10}\selectfont
  \centering
  \begin{tabular}{llccc}
    \toprule
    
    Scene & Method & CLIP Score$\uparrow$ & Consistency$\uparrow$ & Time$\downarrow$ \\
    
    \midrule
    
    \multirow{2}{*}{Portrait}
        & Ours & \textbf{27.75} & \textbf{0.953} & \textbf{60 mins} \\
        & IN4D~\cite{instruct_4d24d} & 27.38      & 0.933       & 2 hours \\

    \midrule

    \multirow{2}{*}{Cat}
        & Ours  & \textbf{31.81}  & \textbf{0.968} & \textbf{60 mins} \\
        & IN4D  & 31.72 & 0.964 & 2 hours \\

    \midrule

    \multirow{2}{*}{Steak}
        & Ours  & \textbf{28.52}  & \textbf{0.988} & \textbf{40 mins} \\
        & IN4D  & 28.23 & 0.983 & 2 hours \\
        
    \bottomrule
  \end{tabular}
  \caption{Quantitative comparison with Instruct 4D-to-4D (IN4D). The results demonstrate that our method outperforms IN4D in better alignment with text prompts, superior temporal consistency, and faster runtime.}
  \label{tab:comparison}
\end{table}

\vspace{-0.5cm}
\paragraph{Comparison with video editing methods.}
We further compare our method with existing video editing approaches on monocular scenes. We use AnyV2V~\cite{ku2024anyv2v} as our baseline, editing only the first frame and then propagating changes to the remaining frames, which is similar to ours. As shown in Fig.~\ref{fig:comparison_video}, we edit the first frame (leftmost) and generate our results directly using the personalized IP2P model, without dynamic scene optimization. Our method produces significantly higher quality and more consistent results, whereas AnyV2V outputs are blurry and display inconsistencies in the face and suit across frames. Furthermore, as the input video length increases, AnyV2V becomes increasingly unstable and produces lower-quality results. In contrast, our method edits frame-by-frame, handling inputs of any length. Additionally, our method supports both monocular and multi-camera scenes, while AnyV2V struggles to maintain consistency across multiple camera views.

\subsection{Quantitative Evaluation}
\vspace{-0.3cm}
We present a quantitative comparison with Instruct 4D-to-4D (IN4D)~\cite{instruct_4d24d} in Tab.~\ref{tab:comparison} on monocular scenes (Portrait, Cat) and a multi-camera scene (Steak). We utilize the CLIP score to measure the alignment of the editing results with text instructions. To evaluate temporal consistency, we use a subject consistency score (denoted as \textbf{Consistency}) from VBench~\cite{huang2023vbench}, which is designed to evaluate subject consistency in video generative models. Specifically, the consistency score is computed by measuring the DINO~\cite{caron2021emerging} feature similarity across frames. Our results achieve higher CLIP and consistency scores, indicating better alignment with the text prompts and superior temporal consistency. Additionally, we compare the running time of each method. The total time required for our method, including fine-tuning the IP2P model and optimizing dynamic scenes, is less than half that of IN4D. More details are discussed in the appendix.

\begin{figure}[t]
  \centering
  \includegraphics[clip, trim=0cm 1.8cm 8.3cm 0cm, width=1.0\linewidth]{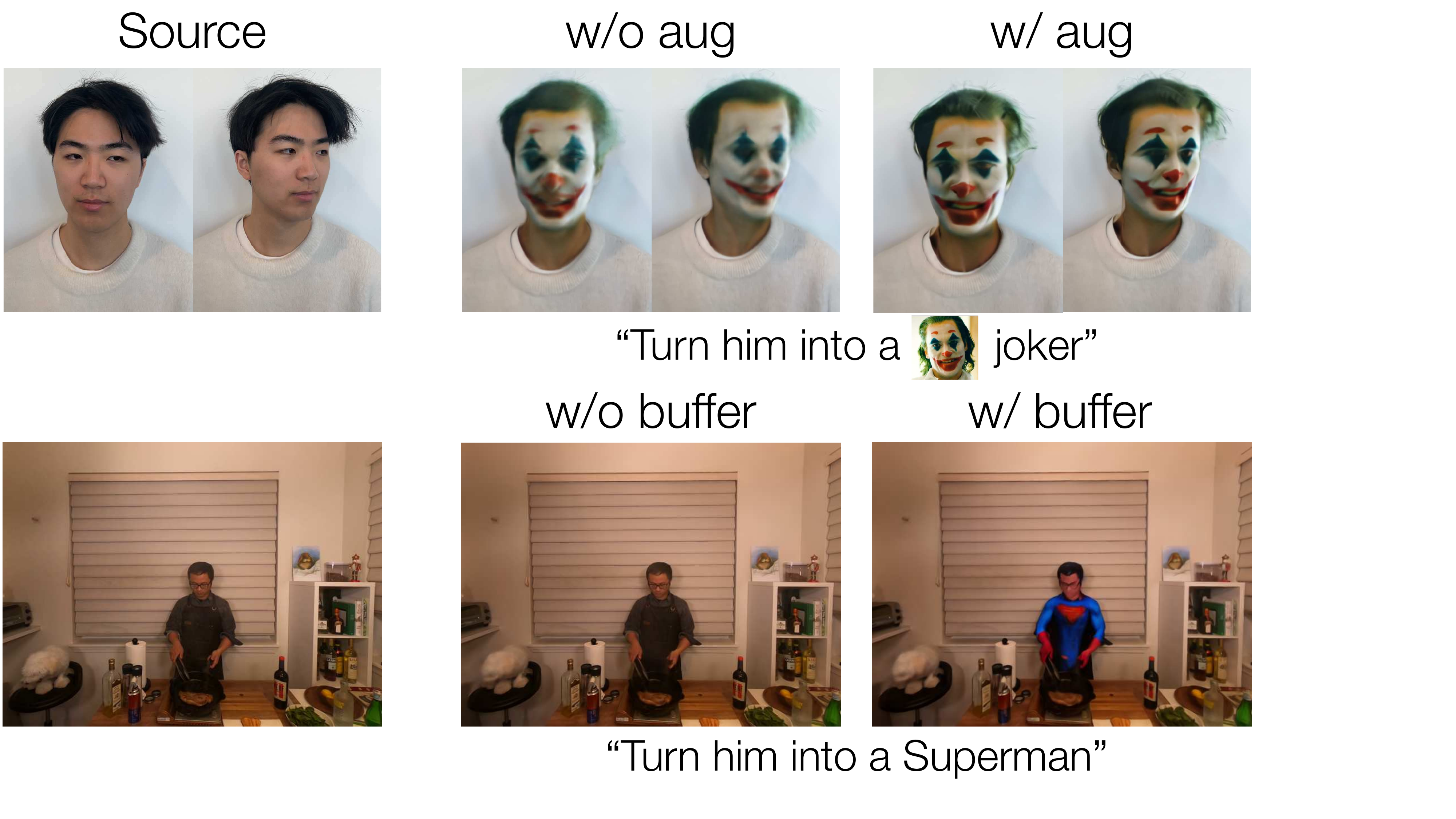}
  \vspace{-5mm}
  \caption{Ablation study. The results demonstrate the effectiveness of our data augmentation and improved efficiency of using our edited image buffer.}
  \label{fig:ablation_study}
\end{figure}

\subsection{Ablation Study}

\paragraph{Data augmentation.} We evaluate the impact of data augmentation in IP2P fine-tuning in the upper part of Fig.~\ref{fig:ablation_study}. We fine-tune IP2P without data augmentation (denoted as \textbf{w/o aug}) and proceed with dynamic scene optimization. The results are blurry and fail to produce smooth, consistent editing across frames due to the inconsistencies in the frames generated by the personalized IP2P. In contrast, fine-tuning IP2P with data augmentation (denoted as \textbf{w/~aug}) yields significantly higher quality and more consistent results.

\vspace{-0.5cm}
\paragraph{Edited image buffer.} We evaluate the effectiveness of our designed edited image buffer for optimizing dynamic 3D Gaussians. For comparison, we introduce a variant that performs optimization by randomly selecting a frame from all frames to optimize the Gaussians (denoted as \textbf{w/o buffer}). In contrast, our method specifically selects frames from the edited image buffer (denoted as \textbf{w/ buffer}). We run both methods for $1000$ optimization iterations. As shown in the lower part of Fig.~\ref{fig:ablation_study}, our method successfully produces the edited results, whereas the variant generates results nearly identical to the original scene, highlighting the importance of the edited image buffer.

\begin{figure}[t]
  \centering
  \includegraphics[clip, trim=0cm 14.2cm 0cm 0cm, width=1.0\linewidth]{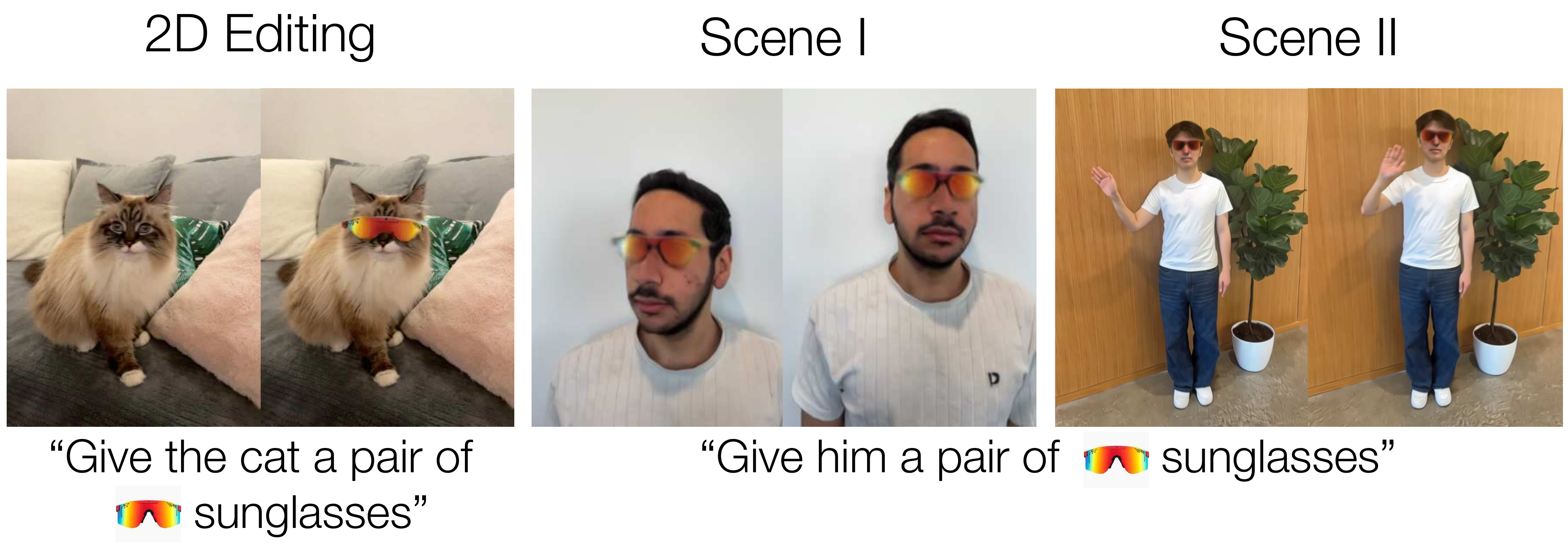}
  \caption{Example of editing ability generalization. We demonstrate that by fine-tuning a personalized IP2P model using images of a cat, it can be successfully applied to other domains, such as portrait and full-body scenes.}
  \label{fig:transfer}
\end{figure}

\subsection{Editing Ability Generalization}

Since our personalization resembles IP2P ``learning'' a new editing ability based on the reference image, we validate its ability to generalize this ability to data outside the reference domain. The results are shown in Fig.~\ref{fig:transfer}. In this example, we begin by applying image-driven editing on a 2D image to add specific sunglasses to a cat. This edited image is then used as a reference to fine-tune IP2P, which subsequently applies the learned editing ability to a portrait scene and a full-body scene. The results are stable and high-fidelity, demonstrating the robust generalization capabilities of our personalized IP2P.

\begin{figure}[t]
  \centering
  \includegraphics[clip, trim=0cm 4.2cm 5.5cm 0cm, width=1.0\linewidth]{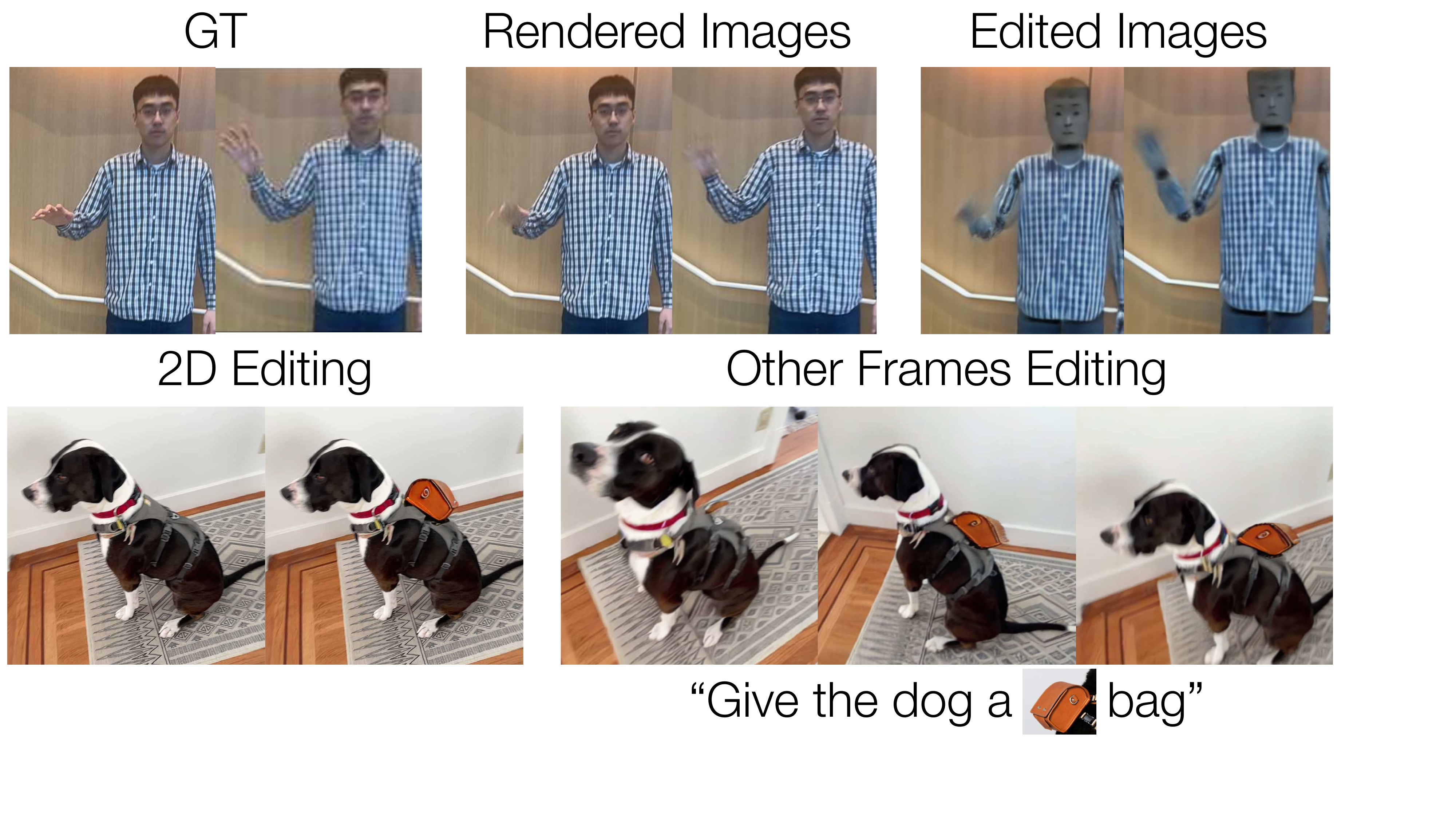}
  \caption{Examples of failure cases. The reconstruction backbone fails to produce clear rendering results, leading to blurry edited images (top). Our method exhibits limitations inherited from IP2P, such as adding a specific bag within empty regions (bottom).}
  \label{fig:limitation}
\end{figure}

\subsection{Limitation}

While our method produces high-quality editing results, it inherits certain limitations from both our reconstruction backbone and IP2P. As shown in the first row of Fig.~\ref{fig:limitation}, when the input trained dynamic 3D Gaussians scene fails to render clear images, especially the moving hands, our framework produces edits based on these unclear images, which results in blurry and unclear outcomes. Moreover, while our fine-tuning significantly alleviates the original IP2P’s difficulty with controllable 2D edits, our personalized IP2P still generates 3D-inconsistent images when editing in empty regions. In the second row of Fig.~\ref{fig:limitation}, we give the dog a specific bag, but the model struggles to consistently render the bag from multiple views. This issue likely arises from IP2P’s difficulty in adding complex content to empty regions. Future improvements to our method could benefit from a more robust reconstruction backbone and a base diffusion model with enhanced capabilities for adding content. Further limitations are discussed in the appendix.

\section{Conclusion}
\label{sec:conclusion}

In this paper, we propose \coolname, a novel framework for dynamic 3D scene editing by simplifying the task into 2D editing through fine-tuning InstructPix2Pix with a single pair of images, combined with an efficient two-stage optimization of deformable 3D Gaussians. Our approach is compatible with any 2D editing tool, enabling controllable and flexible local editing in dynamic scenes. Qualitative results demonstrate high-quality, high-fidelity, and consistent edits, while our comparisons show that our method significantly outperforms baselines, underscoring its effectiveness. We envision our framework advancing controllable scene editing and inspiring future exploration in this field.

{
    \small
    \bibliographystyle{ieeenat_fullname}
    \bibliography{bib/2d_editing, bib/3d_editing, bib/4d_editing, bib/dynamic}
}

\end{document}